%% file: egpaper_for_review.tex
\documentclass[10pt,twocolumn,letterpaper]{article}

\usepackage{iccv}
\usepackage{times}
\usepackage{epsfig}
\usepackage{graphicx}
\usepackage{amsmath}
\usepackage{amssymb}
\usepackage{booktabs}
\usepackage{multirow,makecell, rotating}
\usepackage{comment}
\usepackage{bbm}
\usepackage{xcolor}
\usepackage{fancyhdr}
\usepackage{setspace}
\usepackage{graphbox}
\usepackage[accsupp]{axessibility}  

\usepackage[pagebackref=true,breaklinks=true,letterpaper=true,colorlinks,bookmarks=false]{hyperref}

\usepackage[capitalize]{cleveref}
\crefname{section}{Sec.}{Secs.}
\Crefname{section}{Section}{Sections}
\Crefname{table}{Table}{Tables}
\crefname{table}{Tab.}{Tabs.}

\usepackage[acronym]{glossaries}
\newacronym{Task}{VSD}{Visual similarity discovery}

\iccvfinalcopy 

\def\iccvPaperID{11744} 

\ificcvfinal\fi

\begin{document}

\title{Efficient Discovery and Effective Evaluation of Visual Perceptual Similarity:\\ A Benchmark and Beyond}

\author{\vspace{2mm} Oren Barkan\thanks{\hspace{0.2mm}Equal contribution.}$^{\hspace{1.5mm}1}$
\hspace{10mm} 
Tal Reiss$^{\ast \hspace{0.2mm} 2}$
\hspace{10mm}
Jonathan Weill$^{3}$ \\
\vspace{3mm}
Ori Katz$^{4}$
\hspace{10mm}
Roy Hirsch$^{3}$ 
\hspace{10mm}
Itzik Malkiel$^{3}$ 
\hspace{10mm}
Noam Koenigstein$^{3}$ 
\\
\vspace{2mm}
\normalsize{$^{1}$The Open University}
\hspace{7mm} \normalsize{$^{2}$The Hebrew University of Jerusalem}
\hspace{7mm} \normalsize{$^{3}$Tel Aviv University}
\hspace{7mm} \normalsize{$^{4}$Technion}
\\
\normalsize{\url{https://vsd-benchmark.github.io/vsd}}
}
\maketitle
\ificcvfinal\thispagestyle{empty}\fi

\newcommand*\samethanks[1][\value{footnote}]{\footnotemark[#1]}
\newcommand\blfootnote[1]{%
  \begingroup
  \renewcommand\thefootnote{}\footnote{#1}%
  \addtocounter{footnote}{-1}%
  \endgroup
}

\begin{abstract}
\input{01_abstract.tex}
\end{abstract}

\section{Introduction}
\label{sec:intro}
\input{02_intro.tex}

\section{Related work}
\label{sec:related}
\input{03_related.tex}

\input{eds.tex}

\begin{table}[t]
  \begin{center}
  \begin{tabular}{lcc}
    \toprule
Benchmark & \# of queries &	\# of annotated  pairs	\\
\cmidrule{1-1}
\cmidrule{2-2}
\cmidrule{3-3}
Discovery & 2,000 & 54,170 \\
Wild & 2,000 & 64,046 \\
    \bottomrule
  \end{tabular}
  \end{center}\vspace{-2mm}
  \caption{Ground-truth statistics.}\vspace{-3mm}
    \label{tab:stats}
\end{table}

\section{\gls{Task} benchmarks}
\label{sec:benchmark}
The proposed benchmarks are based on the DeepFashion (DF) \cite{deepfashion} dataset. DF contains over $800,000$ images of fashion items. These images include both "in-shop" images that display clothing items worn by models in a clean environment, as well as images taken by consumers in the wild. 
The benchmarks utilize two subsets of the DF dataset. The first subset, named In-shop Clothes Retrieval (ICR), contains $52,712$ images of clothes worn by models. ICR images exhibit a variety of poses and scales. The second, named Consumer-to-shop Clothes Retrieval (CCR), contains $239,557$ images of clothes taken by consumers (wild images). Accordingly, we propose two benchmarks: (i) Closed-catalog discovery. (ii) Image in the wild discovery. 

\begin{figure}[t]
\centering
\begin{tabular}{@{\hskip2pt}c@{\hskip2pt}c@{\hskip2pt}@{\hskip2pt}c@{\hskip2pt}c@{\hskip2pt}c@{\hskip2pt}c@{\hskip2pt}c}
& Query & Candidate\\


\\
\rotatebox[origin=c]{90}{Positive} &
\includegraphics[align=c, width=0.46\linewidth]{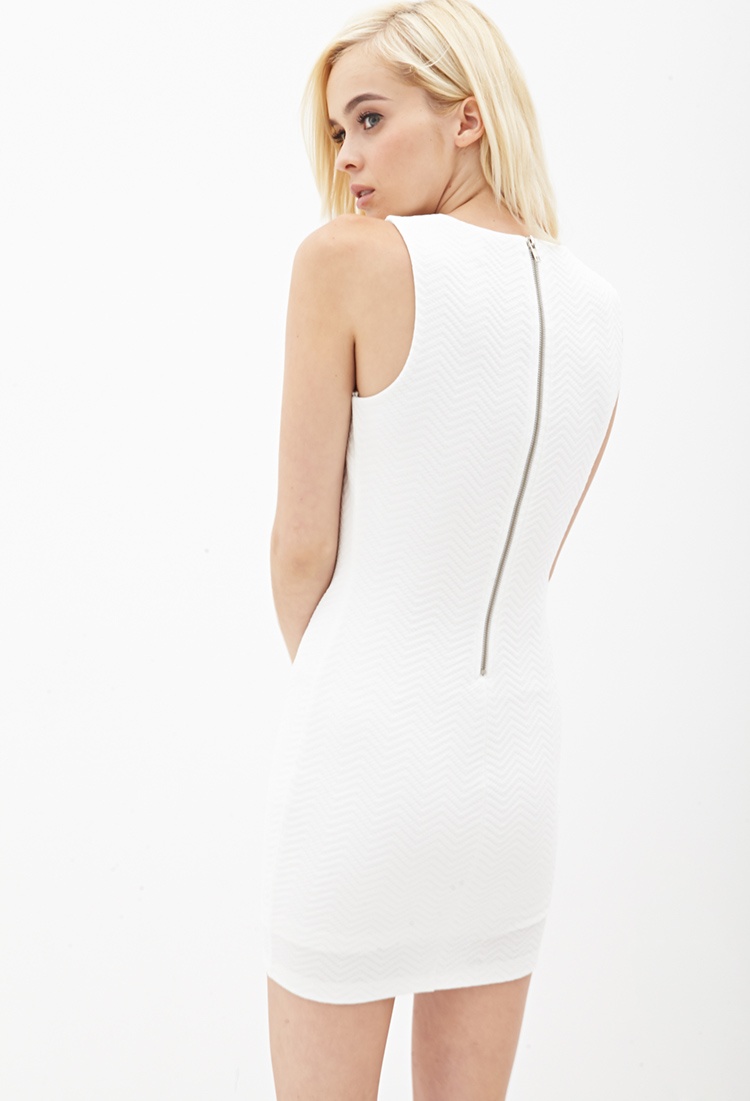} & 
\includegraphics[align=c, width=0.46\linewidth]{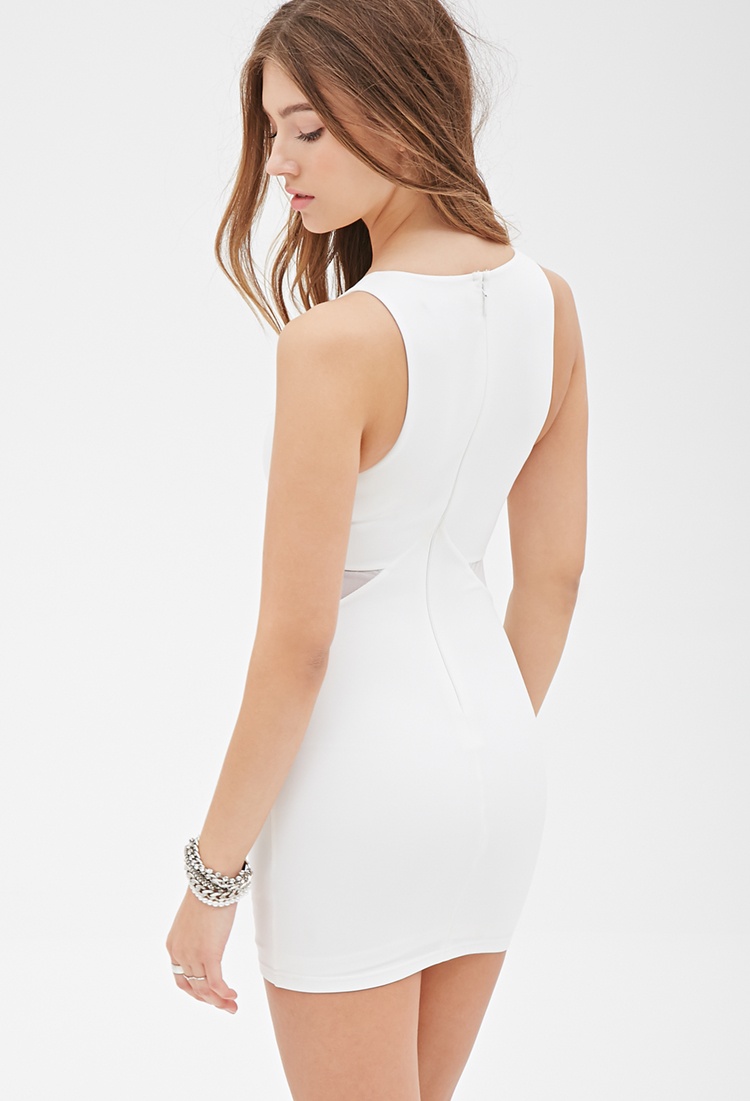} 
\\ 
\\
\rotatebox[origin=c]{90}{Negative} &
\includegraphics[align=c, width=0.46\linewidth]{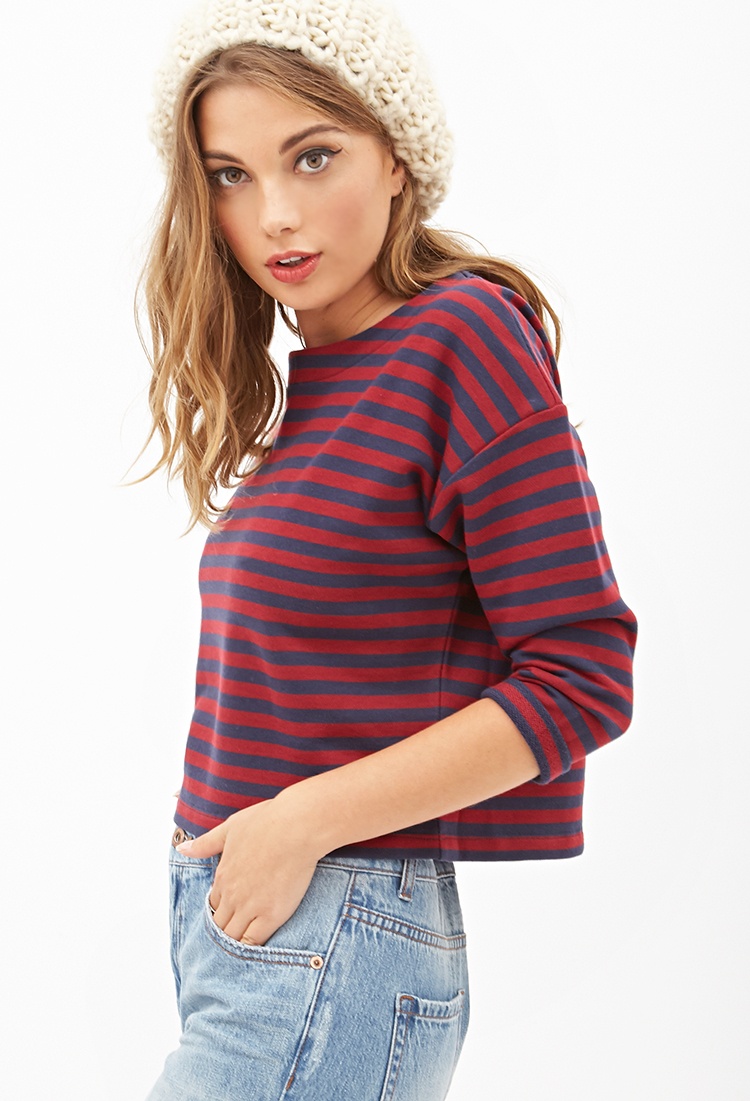} & 
\includegraphics[align=c, width=0.46\linewidth]{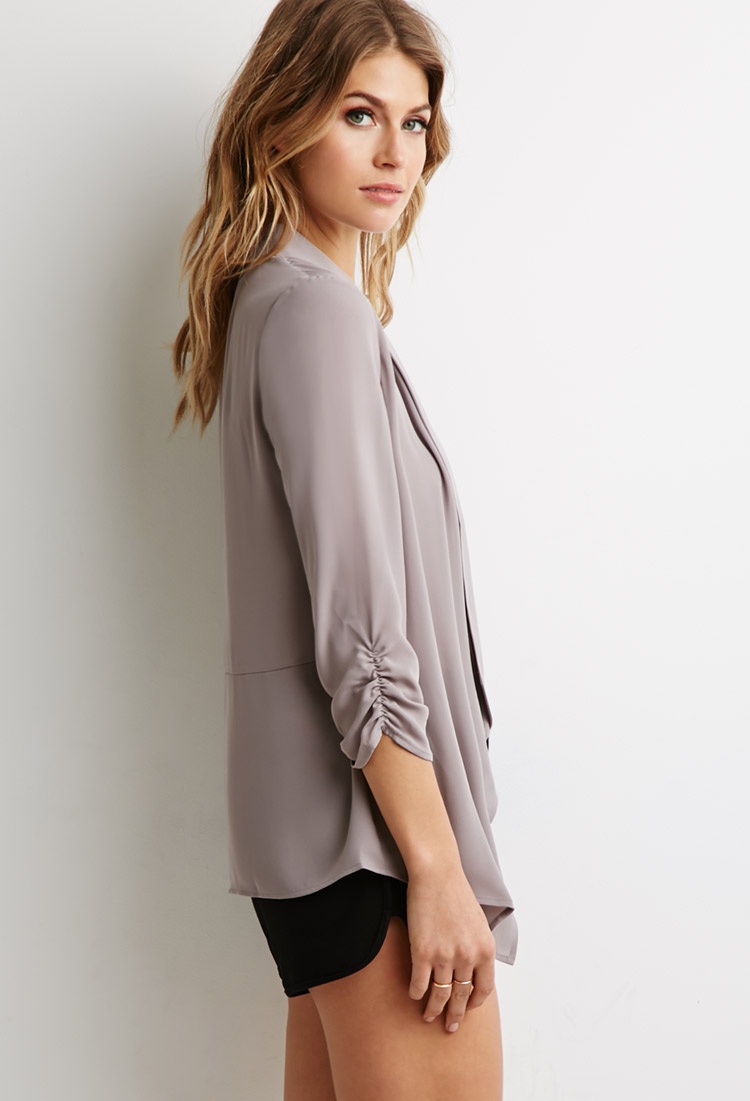}  \\ 
\end{tabular}
\vspace{1em}
\caption{Closed catalog query-candidate pairs. A positive query-candidate pair is presented in the upper row, while a negative query-candidate pair is presented in the lower row.}
\label{fig:discovery_qualitative}
\end{figure}

\subsection{Closed-catalog discovery}
\label{benchmark:closed_catalog}
In the \textit{Closed-catalog} discovery benchmark, we build upon the ICR dataset, where the query images are taken from the catalog. The task is to retrieve images associated with \textit{different} objects, which are similar to the item in the query image. In the Closed-catalog discovery benchmark, $2,000$ query images were selected (denoted by $\mathcal{Q}$), and $|M|=6$ \gls{Task} models were used throughout the EDS procedure (more details about the vision models can be found in \cref{exp:baselines}). Specifically, $\forall q\in \mathcal{Q}$, and $m\in M$ we construct the set $\mathcal{S}_k^m$ (as in \cref{eq:skm}) using $k=6$. Our final set $\mathcal{S}_k=\cup_{m\in M}\mathcal{S}_k^m$, comprised of $54,170$ \emph{query-candidate} image pairs, was evaluated by human domain experts that generated the ground truth (GT) labels of this benchmark.

\begin{figure}[t]
\centering
\begin{tabular}{@{\hskip2pt}c@{\hskip2pt}c@{\hskip2pt}@{\hskip2pt}c@{\hskip2pt}c@{\hskip2pt}c@{\hskip2pt}c@{\hskip2pt}c}
& Query & Candidate\\


\\
\rotatebox[origin=c]{90}{Positive} &
\includegraphics[align=c, width=0.34\linewidth]{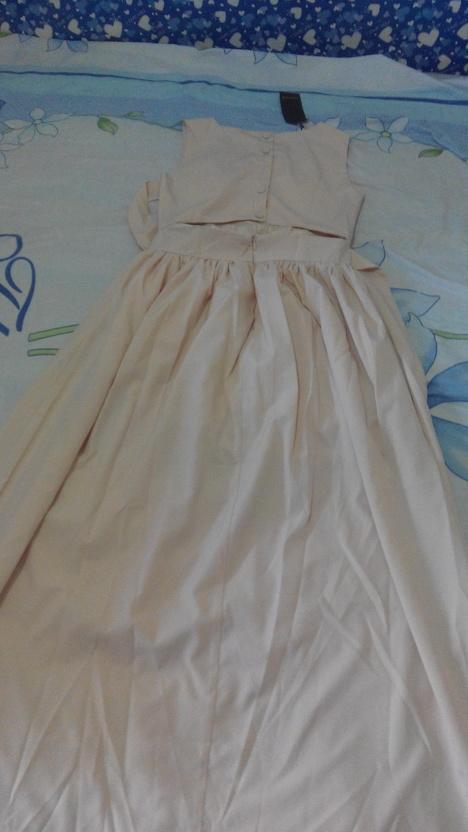} & 
\includegraphics[align=c, width=0.46\linewidth]{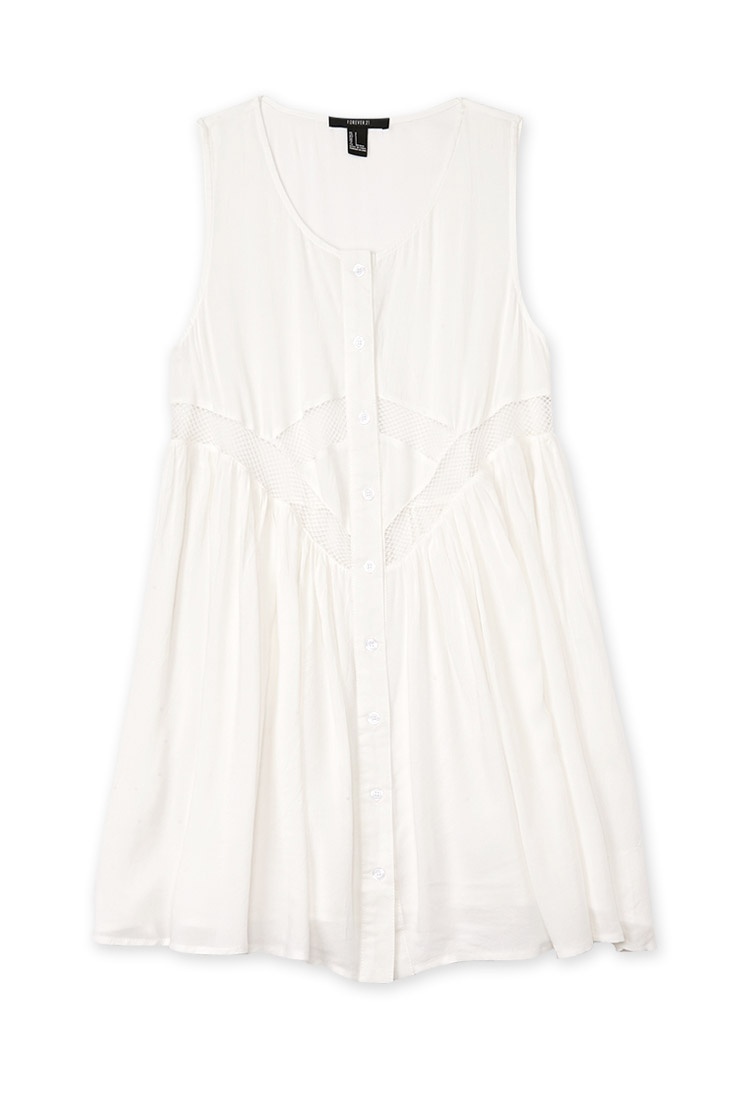} 
\\ 
\\
\rotatebox[origin=c]{90}{Negative} &
\includegraphics[align=c, width=0.46\linewidth]{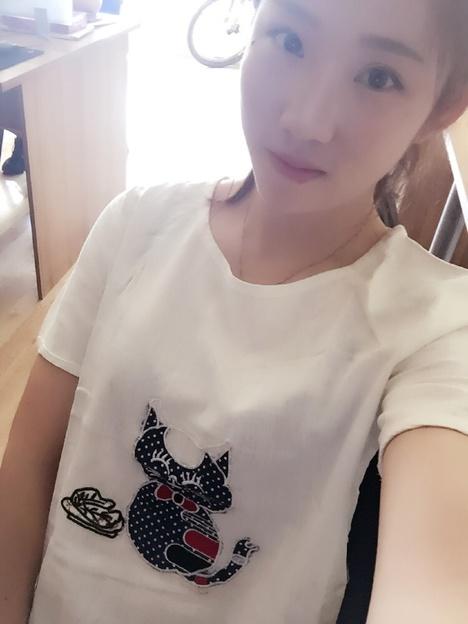} & 
\includegraphics[align=c, width=0.46\linewidth]{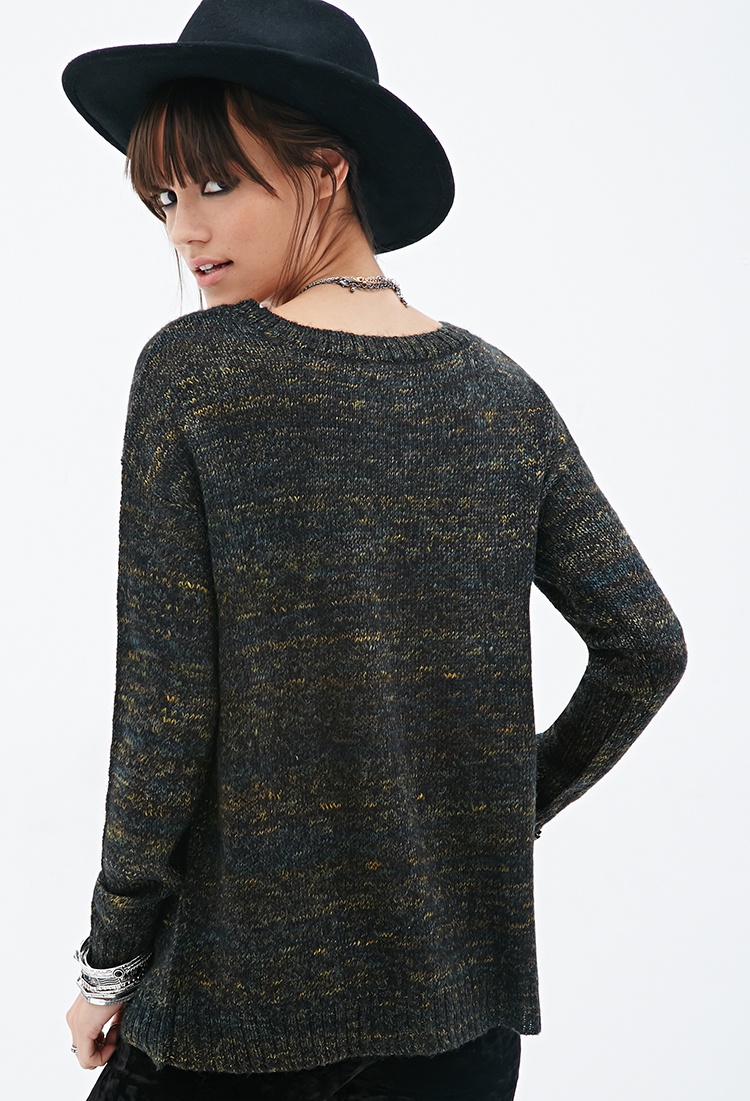}  \\ 
\end{tabular}
\vspace{1em}
\caption{Image in the wild query-candidate pairs. A positive query-candidate pair is presented in the upper row, while a negative query-candidate pair is presented in the lower row.}
\label{fig:wild_qualitative}
\end{figure}

\subsection{Image in the wild discovery}
\label{benchmark:wild}
This benchmark focuses on the discovery of perceptually similar objects where the query images were taken \textit{in the wild}. In this setting, we are given a ``wild'' query image of a clothing item and the task is to retrieve images of perceptually similar items from the ICR dataset. To this end, we adopt wild images from the Consumer-to-shop Clothes Retrieval (CCR) Benchmark dataset and match them with candidates from the ICR dataset. Note that both datasets are associated with significantly different distributions since the ICR dataset contains "in-shop" images and the CCR dataset incorporates images in the wild. 
A total of $2,000$ query images from the CCR dataset were filtered, and the EDS procedure was applied to generate ground truth annotations of image pairs, consisting of query-candidate image pairs from the CCR and ICR datasets, respectively. In total, this benchmark consists of $64,046$ \emph{query-candidate} image pairs. Since objects are often seen with suboptimal lighting, angles, resolution, or cluttered indoor backgrounds, this benchmark presents additional visual challenges compared to Closed-catalog discovery. Importantly, due to the significantly different characteristics of the query and candidate images, this benchmark imposes a cross-domain \gls{Task} task which can emphasize the generalization and robustness of the underlying tested models.

\section{Experiments}
This section answers the following research questions:
\begin{itemize}
  \item RQ1: How effective and fair is our evaluation?
  \item RQ2: Does labeled supervised training improve \gls{Task}?
  \item RQ3: Recognition vs. discovery. How do they differ?
\end{itemize}
We will begin by describing the experimental settings, followed by answering the above three research questions.

\subsection{Baselines}

\label{exp:baselines}
Our pretrained baselines set $M$ was assembled by powerful and successful image classification models from the past few years. Furthermore, the pretraining schemes that the models were pretrained on are highly diverse, resulting in a wide range of predictions. Specifically, we adopt: \emph{Argus (AS)} an open source ResNext101 $32\times8$d pretrained on Bing web data\footnote{\url{https://pypi.org/project/argusvision/}}, \emph{DINO} \cite{dino},  self-supervised pre-trained on ImageNet1K, \emph{BEiT} \cite{beit}, pretrained on ImageNet21K \cite{imagenet21k} and \emph{CLIP} \cite{clip} image encoder, pretrained on web-scale data.

We further consider finetuned versions of the pretrained backbones, where the finetuning is performed using  (i) identity labels, i.e. the gallery item's ID (denoted by ID), (ii) category labels, i.e., the gallery item's category (denoted by CAT), or both. Accordingly, a finetuned version of model X is denoted by X-ID, X-CAT, or X-ID+C (if both ID and CAT information sources were used to finetune X). Specifically, the optimization w.r.t. to each information source (ID and/or CAT) is performed by placing a new linear classification head (on top of the backbone) that matches the output dimension (number of IDs or categories), and the model's parameters are optimized w.r.t. the categorical cross-entropy loss. The exact optimization details are provided in the supplementary material (SM).

\subsection{Candidates diversity}
In \cref{tab:statsistics}, we present the candidates overlap for each pair of GT generators. This analysis demonstrates the diversity of candidates produced by the different GT generators, with an average pairwise overlap of $\sim$10\% (average of the numbers in \cref{tab:statsistics}). Moreover, the maximum number of distinct candidates per query is 30 (as each GT generator can contribute its top-5 candidates at most). We found that after candidate deduping, the average number of candidates per query is 24.4 (a relatively low duplication rate of $\sim$18\%).

\begin{table}[t]
  \begin{center}
  \begin{tabular}{l|cccccc}
    \hline
& AS & DINO & BEiT & CLIP & ID & ID+C \\
\cline{1-7}
AS & - & 15.0 & 16.5 & 11.2 & 3.8 & 3.9 \\
DINO & 15.0 & - & 27.6 & 13.2 & 4.8 & 5.1 \\
BEiT & 16.5 & 27.6 & - & 17.7 & 5.8 & 6.2 \\
CLIP & 11.2 & 13.2 & 17.7 & - & 4.3 & 4.6 \\

ID & 3.8 & 4.8 & 5.8 & 4.3 & - & 15.8 \\
ID+C & 3.9 & 5.1 & 6.2 & 4.6 & 15.8 & - \\
\hline
  \end{tabular}
    \end{center}
    \caption{Pairwise overlap (\%) between GT generator candidates in the closed-catalog task. ID and ID+C have been finetuned with the Argus backbone.}
    \label{tab:statsistics}
\end{table}

\begin{table*}[t]
\begin{minipage}{0.55\textwidth}
  \begin{center}
  \begin{tabular}{lcccccc}
    \toprule
 \multirow{2}{*}{Metric}	& \multicolumn{3}{c}{Micro ROC-AUC} 	&	\multicolumn{3}{c}{Macro ROC-AUC} \\
\cmidrule(r){2-4}
\cmidrule(r){5-7}
	&	AUC 	&	SC & $p$-value	 & AUC & SC & $p$-value \\
\cmidrule(r){1-1}
\cmidrule(r){2-4}
\cmidrule(r){5-7}
AS & 62.6±1.8 & 0.83 & 0.042 & 78.5±1.7 & 0.75 & 0.083 \\
DINO &  69.9±1.9 & 0.94 & 0.004 & 82.9±1.1 & 0.93 & 0.007 \\
BEiT & \textbf{75.2}±0.9 & 0.94 & 0.004 & \textbf{86.0}±0.8 & 0.93  & 0.007  \\
CLIP & 67.7±1.8 & 0.94 & 0.004 & 80.9±1.7 & 0.93 &  0.007\\
AS ID &  62.2±1.8 &  0.83 & 0.042 & 78.7±1.5 & 0.70 & 0.124 \\
AS ID+C & 65.4±1.2 & 1.00 & 0.000 & 80.5±0.9 & 0.93 & 0.007 \\
    \bottomrule
  \end{tabular}
    \end{center}
  \caption{Leave-one-out robustness ROC-AUC closed-catalog discovery results. The $p$-value is computed with respect to a null hypothesis of zero SC.}
\label{tab:auc_robustness}
\end{minipage}
\begin{minipage}{0.5\textwidth}
  \begin{center}
  \begin{tabular}{lcccc}
    \toprule
\multirow{2}{*}{Metric}	& \multicolumn{2}{c}{HR} 	&	\multicolumn{2}{c}{MRR} \\
\cmidrule(r){2-3}
\cmidrule(r){4-5}
	&	@5 	&	@9 	&	@5     &   @9 \\
\cmidrule(r){1-1}
\cmidrule(r){2-3}
\cmidrule(r){4-5}
AS & 15.2 & 10.9 & 20.5 & 17.6 \\
DINO &  26.0 & 18.3 & 34.5 & 29.5 \\
BEiT & 37.2 & 26.4 & 47.3 & 40.7  \\
CLIP & 25.2 & 18.1 & 32.7 & 28.2 \\
AS ID &  \textbf{99.9} & \textbf{99.9} & \textbf{91.8} & \textbf{83.2} \\
AS ID+C & 99.8 & \textbf{99.9} & 89.4 & 80.8 \\
    \bottomrule
  \end{tabular}
  \end{center}
\caption{Identity recognition performance. Bold is best.}
  \label{tab:fingerpringting}
  \end{minipage}
\end{table*}

\begin{table*}[t]
  \begin{center}
    \resizebox{0.95\linewidth}{!}{%
  \begin{tabular}{llcccccccccccc}
    \toprule
\multirow{3}{*}{} &  &  & 	&	&  &	\multicolumn{4}{c}{ROC-AUC} & \multicolumn{4}{c}{PR-AUC} 	\\
\cmidrule(r){7-10}
\cmidrule(r){11-14}
& 	&	\multicolumn{2}{c}{HR} 	& 	\multicolumn{2}{c}{MRR}  &   \multicolumn{2}{c}{Micro} &  \multicolumn{2}{c}{Macro} &  \multicolumn{2}{c}{Micro} &  \multicolumn{2}{c}{Macro} \\
\cmidrule(r){3-4}
\cmidrule(r){5-6}
\cmidrule(r){7-8}
\cmidrule(r){9-10}
\cmidrule(r){11-12}
\cmidrule(r){13-14}
& Method	&	@5 	&	@9 	&	@5     &   @9 & Anno. & Neg & Anno. & Neg & Anno. & Neg & Anno. & Neg\\
\cmidrule(r){2-2}
\cmidrule(r){3-3}
\cmidrule(r){4-4}
\cmidrule(r){5-5}
\cmidrule(r){6-6}
\cmidrule(r){7-7}
\cmidrule(r){8-8}
\cmidrule(r){9-9}
\cmidrule(r){10-10}
\cmidrule(r){11-11}
\cmidrule(r){12-12}
\cmidrule(r){13-13}
\cmidrule(r){14-14}
\multirow{6}{*}{\rotatebox{90}{GT Generators}} & Argus & 82.5 & 52.9 & 84.5 & 71.3 & 62.7 & 73.9 & 77.1 & 81.9 & 91.7 & 94.1 & 92.2 & 95.6 \\
& DINO &  91.9 & \textbf{64.0} & 93.1 & \textbf{80.9} & 70.4 & 76.8 & 82.2 & 86.8 & 94.3 & 94.9 & 94.2 & 97.0\\
& BEiT & \textbf{92.3} & 63.2 & \textbf{93.4} & 80.6 & \textbf{75.5} & \textbf{81.6} & \textbf{85.4} & \textbf{88.8} & \textbf{95.4} & \textbf{96.1} & \textbf{95.1} & \textbf{97.4} \\
& CLIP & 81.5 & 52.7 & 84.2 & 71.4 & 67.8 & 74.1 & 79.7 & 83.1 & 93.2 & 94.3 & 92.9 & 95.9 \\
& Argus ID &  84.3 & 51.6 & 86.0 & 71.5 & 62.3 & 76.8 & 77.1 & 83.4 & 91.9 & 95.2 & 92.5 & 96.0\\
& Argus ID+C & 86.8 & 52.5 & 88.9 & 73.6 & 65.4 & 76.6 & 79.1 & 83.0 & 92.7 & 95.1 & 93.3 & 95.8\\
\cline{1-2}
\multirow{8}{*}{\rotatebox{90}{New models}}& Argus CAT & 9.0 & 7.23 & 11.0 & 9.9 & 67.5 & 65.9 & 79.9 & 83.0 & 92.7 & 95.1 & 93.3 & 95.8 \\
& DINO CAT & 4.3 & 3.6 & 4.7 & 4.3 & 71.0 & 68.3 & 81.6 & 76.7 & 94.0 & 91.6 & 94.1 & 93.4 \\
& BEiT CAT & 38.7 & 28.9 & 47.0 & 41.2 & 72.9 & 71.9 & 83.7 & 83.4 & 94.7 & 95.7 & 94.9 & 96.0 \\
& DINO ID & 32.3 & 25.7 & 38.6 & 34.4 & 68.8 & 83.3 & 81.7 & 87.6 & 93.7 & 96.2 & 94.4 & 96.9 \\
& BEiT ID & 34.9 & 27.0 & 41.8 & 37.1 & 71.0 & 81.1 & 82.9 & 88.0 & 94.3 & 96.5 & 94.9 & 97.0 \\
& DINO ID+C & 32.4 & 25.0 & 39.2 & 34.7 & 71.9 & 80.7 & 83.1 & 86.2 & 94.6 & 96.2 & 94.4 & 96.9 \\
& BEiT ID+C & 34.6 & 26.3 & 42.1 & 37.1 & 72.3 & 80.6 & 83.3 & 86.0 & 94.7 & 95.7 & 95.0 & 96.5 \\
& DINO FT & 88.2 & 58.5 & 90.0 & 76.9 & 69.2 & 80.5 & 81.1 & 86.6 & 93.5 & 95.8 & 93.5 & 96.9 \\
\bottomrule
  \end{tabular}
  }
  \end{center}
  \caption{Discovery performance. Bold denotes the best results.}
    \label{tab:models}
    \vspace{-1mm}
\end{table*}

\subsection{Evaluation metrics (RQ1)}
\textbf{Hit Ratio at $k$ (HR@$k$)} \cite{hr}. HR@$k$ is the percentage of the predictions made by the model, where the true item was found in the top $k$ items suggested by the model. A query-candidate example is scored $1$ if the candidate item is ranked among the top $k$ predictions of the model, otherwise $0$. This is followed by the average of all query-candidate examples in the test set. 

\textbf{Mean Reciprocal Rank at $k$ (MRR@$k$)} \cite{mrr}. This measure is defined as the average of the reciprocal ranks considering the top $k$ prediction ranked items. The main difference between MRR and HR is that the former takes into account the order in which predictions are made.

\textbf{The area under the receiver operating characteristic curve (ROC-AUC)} \cite{roc}. It is the metric proposed in Sec.~\ref{sec:effective-eval}. ROC-AUC measures how well the model distinguishes positive from negative data. By varying a threshold, the ROC curve plots true positive rates against false positive rates.

\textbf{The area under the precision-recall curve (PR-AUC)} \cite{pr}. Similar to ROC-AUC, PR-AUC is a threshold-independent metric that calculates the area under a curve. In this case, the curve is defined by a trade-off between precision and recall (the precision as a function of the recall).

It is important to note that we differentiate between two types of AUC: (i) Micro-averaged AUC, computing all query-candidates scores  (ii) Macro-averaged AUC, computing the query-level AUC, then averaging the AUC results over all queries (same as in \cref{eq:avg_auc}).

\subsubsection{AUC for bias reduction}
For the purpose of understanding how robust the AUC metric is to the choice of the annotated set, a leave-one-out experiment was conducted. By using a subset of the GT that does not include predictions from a single model $m\in M$ at a time, the leave-one-out experiment evaluates all the baselines in $M$. As an example, in the first evaluation, the GT was a subset containing all the predictions from our baselines except for Argus. This allows us to see how a model is influenced by the removal of its predictions from the GT. In \cref{tab:auc_robustness} we present the ROC-AUC average micro and macro results over all possible subsets, as well as the spearman correlation (SC) results of each subset of the GT with the full set of pairs. The first column in \cref{tab:auc_robustness} indicates the relevant subsets that were removed. The SC results show a strong correspondence between each subset of the GT and the full set of the GT, emphasizing the robustness of the AUC metric. Additionally, the hierarchy of the models remains the same as in \cref{tab:models}, indicating the AUC robustness.

\begin{table*}[t]
  \begin{center}
  \resizebox{0.95\linewidth}{!}{%
  \begin{tabular}{llcccccccccccc}
    \toprule
\multirow{3}{*}{} &  &  & 	&	&  &	\multicolumn{4}{c}{ROC-AUC} & \multicolumn{4}{c}{PR-AUC} 	\\
\cmidrule(r){7-10}
\cmidrule(r){11-14}
& 	&	\multicolumn{2}{c}{HR} 	& 	\multicolumn{2}{c}{MRR}  &   \multicolumn{2}{c}{Micro} &  \multicolumn{2}{c}{Macro} &  \multicolumn{2}{c}{Micro} &  \multicolumn{2}{c}{Macro} \\
\cmidrule(r){3-4}
\cmidrule(r){5-6}
\cmidrule(r){7-8}
\cmidrule(r){9-10}
\cmidrule(r){11-12}
\cmidrule(r){13-14}
& Method	&	@5 	&	@9 	&	@5     &   @9 & Anno. & Neg & Anno. & Neg & Anno. & Neg & Anno. & Neg\\
\cmidrule(r){2-2}
\cmidrule(r){3-3}
\cmidrule(r){4-4}
\cmidrule(r){5-5}
\cmidrule(r){6-6}
\cmidrule(r){7-7}
\cmidrule(r){8-8}
\cmidrule(r){9-9}
\cmidrule(r){10-10}
\cmidrule(r){11-11}
\cmidrule(r){12-12}
\cmidrule(r){13-13}
\cmidrule(r){14-14}
\multirow{6}{*}{\rotatebox{90}{GT Generators}} & Argus & 55.5 & 33.2 & 56.1 & 46.3 & 71.0 & 72.5 & 75.0 & 81.7 & 60.0 & 89.7 & 68.5 & 92.3 \\
& DINO &  54.0 & 33.5 & 55.0 & 45.9 & 70.8 & 72.9 & 75.1 & 84.0 & 60.5 & 89.1 & 68.3 & 93.3\\
& BEiT & \textbf{69.6} & \textbf{43.0} & \textbf{71.0} & \textbf{59.2} & \textbf{77.2} & \textbf{77.7} & \textbf{81.3} & \textbf{87.0} & \textbf{68.5} & \textbf{91.6} & \textbf{75.7} & \textbf{94.7} \\
& CLIP & 12.3 & 9.9 & 14.4 & 13.0 & 59.0 & 60.7 & 67.7 & 74.9 & 50.6 & 84.1 & 63.0 & 89.0 \\
& Argus ID & 8.8 & 5.4 & 8.9 & 7.4 & 38.8 & 54.3 & 41.4 & 66.7 & 33.8 & 81.0 & 44.7 & 84.8 \\
& Argus ID+C & 5.1 & 3.5 & 5.1 & 4.4 & 39.7 & 52.4 & 41.4 & 64.5 & 34.1 & 79.3 & 44.3 & 83.1\\
\cline{1-2}
\multirow{8}{*}{\rotatebox{90}{New models}}& Argus CAT & 0.8 & 0.7 & 0.9 & 0.8 & 46.2 & 53.7 & 49.3 & 65.1 & 39.7 & 79.7 & 50.3 & 84.0 \\
& DINO CAT & 0.7 & 0.6 & 0.8 & 0.8 & 64.8 & 63.4 & 66.2 & 63.4 & 55.6 & 84.2 & 62.9 & 86.5 \\
& BEiT CAT & 14.5 & 11.0 & 18.2 & 16.0 & 72.9 & 72.4 & 75.0 & 80.2 & 64.4 & 88.6 & 71.2 & 91.7 \\
& DINO ID & 5.8 & 4.7 & 6.8 & 6.1 & 61.2 & 70.4 & 64.9 & 75.7 & 52.1 & 88.6 & 62.7 & 90.0 \\
& BEiT ID & 13.7 & 10.6 & 16.3 & 14.5 & 72.4 & 76.4 & 74.4 & 81.6 & 64.1 & 91.1 & 71.2 & 92.4 \\
& DINO ID+C & 5.2 & 4.2 & 5.9 & 5.4 & 62.0 & 69.3 & 65.2 & 75.4 & 53.2 & 88.0 & 63.1 & 89.7 \\
& BEiT ID+C & 15.8 & 12.5 & 18.6 & 16.6 & 70.6 & 77.5 & 73.5 & 82.2 & 62.9 & 91.9 & 70.7 & 92.8 \\
& DINO FT & 29.5 & 22.3 & 34.1 & 30.2 & 71.8 & 76.3 & 73.9 & 83.8 & 62.7 & 91.3 & 68.3 & 93.3 \\
\bottomrule
  \end{tabular}
  }
  \end{center}
      \label{tab:models_wild}
    \caption{Image in the wild performance. Bold denotes the best results.}\vspace{-2mm}
\end{table*}

\subsection{Performance comparison (RQ2)}
We present a comprehensive comparison of all the baselines which includes both HR@$k$ and MRR@$k$ metrics as well as the robust AUC metrics. In addition, we computed the AUC metrics in two different ways: (i) using negatives annotated by our human domain expert annotators, and (ii) using negatives randomly selected from a set of the top $100$ to $500$ predictions of the baseline models. The second scenario assumes that there are only negative predictions after the top $100$ predictions of each model. The comparison also includes the evaluation of GT generator models, various versions of supervised finetuning (ID and/or CAT), and a finetuned DINO with its own self-supervised objective.

In order to demonstrate that even without being a generator, it is possible to obtain good discovery results, in the SM we provide results for \emph{8 additional} models on our annotated GT. We evaluated supervised models: ResNet50 \cite{resnet} and ConvNext \cite{convnet} pretrained on ImageNet1K, ViT B-16 \cite{vit} pretrained on ImageNet21K and SwAG \cite{swag} weakly supervised pretrained through hashtags. Self-supervised models: MoCo \cite{moco}, SwAV \cite{swav}, MAE \cite{mae}, and NoisyStud \cite{noisystud}.

\textbf{Closed-catalog discovery results.} The closed-catalog discovery task results are presented in \cref{tab:models}. We observe bias in HR@$k$ and MRR@$k$ metrics when analyzing models that were not included in the GT generation. Our conclusion is that HR@$k$ and MRR@$k$ are only relevant for the baselines that generated the GT. It is also important to note that there are some inconsistencies between the hierarchies produced by HRR@$k$ and MRR@$k$ and those generated by AUC. In particular, we note that Argus ID and Argus ID+C improve the performance of the Argus backbone when it comes to HR@$k$ and MRR@$k$, but when it comes to AUC, Argus ID actually degrades performance. Our analysis of this behavior revealed that the AUC metrics penalize significantly negative predictions that are top-ranked (i.e., predictions that are found to be negative by our human domain expert annotators), whereas HR@$k$ and MRR@$k$ do not penalize significantly negative predictions. By using a negative sampling strategy for computing AUCs, we find that the hierarchies are closer to those of HR@$k$ and MRR@$k$.

Moreover, finetuning with ID was not found to be helpful, and in fact, degraded the results. However, while finetuning with category labels resulted in a small performance boost for Argus and DINO, it resulted in a performance decrease for BEiT. Considering this, we conclude that there is a great deal of work to be done in this area, and in particular, finding a training scheme that improves the level of similarity. We will, however, leave this for future research.

\textbf{Image in the wild discovery results.} We present the image in the wild results in \cref{tab:models_wild}. In this setting, as in the closed catalog, the HR@$k$ and MRR@$k$ metrics are skewed towards the GT baselines, thereby misevaluating new models. Moreover, we found that supervised finetuning approaches fall significantly behind their pretrained counterparts, as evidenced by the significant degradation in the AUC metrics in each of the backbones. As an example, pretrained DINO results in 70.8\% micro ROC-AUC, while DINO CAT, DINO ID, and DINO ID+C result in 64.8\%, 61.2\%, and 62.0\% micro ROC-AUC, respectively. Nevertheless, the finetuned version of DINO, with its own self-supervised objective, results in a performance improvement of 71.8\% micro ROC-AUC. This may imply that finetuning using supervised objectives results in a loss of generability.

\subsection{Identification is not discovery (RQ3)}
We present the results of the identity recognition task for each baseline in \cref{tab:fingerpringting}. As can be seen, the supervised baselines with identity labels are essentially solving the task, while the pretrained models perform significantly worse. This result highlights the large difference between the identification task and the discovery task. The supervised baselines do not achieve the same level of performance as their other models (e.g. BEiT and DINO) in the discovery task, despite the fact that they solve the task of identification.

\section{Conclusion}
In this paper, we revisit the challenges of evaluating methods for VSD. We demonstrated the limitations of the common practice which is based on identification-retrieval tasks, thereby motivating the need for utilizing domain experts' annotations. We introduced a novel method for efficiently labeling a similarity dataset using human domain experts. We discussed its limitations and proposed evaluation metrics to mitigate them. We employed the proposed method on the DF dataset and curated an annotated dataset consisting of more than 110K image pairs. To the best of our knowledge, this is the first large-scale benchmark for evaluating VSD models in the fashion domain. We hope that our work and the released dataset will expedite VSD research. In the future, we plan to apply the proposed method and metrics for discovery and evaluation of perceptual similarity in other application domains such as natural language understanding~\cite{jiang2019smart,barkan2020scalable,barkan2017bayesian,mueller2016siamese} and audio analytics~\cite{aucouturier2002music,knees2013survey,pampalk2005improvements}.

\section*{Acknowledgements}
We express our sincere appreciation to Lior Greenberg for crafting the project's website, as well as for the fruitful discussions and feedback.

{\small
\bibliographystyle{ieee_fullname}
\bibliography{egbib}
}

\newpage
\clearpage

\appendix
\centerline{\textbf{\LARGE Appendix}}
\vspace{1.5em}
\centerline{\textbf{\Large Efficient Discovery and Effective}}
\centerline{\textbf{\Large Evaluation of Visual Perceptual Similarity:}}
\centerline{\textbf{\Large A Benchmark and Beyond}}
\vspace{1em}

\section{Further evaluation}
\subsection{Additional models}
In order to demonstrate that even without being a generator, it is possible to obtain good discovery results, we provide results for \emph{8 additional} pre-trained models on our annotated GT in \cref{tab:discovery_extra} and \cref{tab:wild_extra}. We evaluated supervised top models: ResNet50 \cite{resnet} and ConvNext (CN) \cite{convnet} pretrained on ImageNet1K, ViT B-16 \cite{vit} pretrained on ImageNet21K and SwAG \cite{swag} weakly supervised pretrained through hashtags. Self-supervised top methods: MoCo \cite{moco}, SwAV \cite{swav}, MAE \cite{mae}, and NoisyStud (NS) \cite{noisystud}.

We observe that the results obtained by the newly added models are on par with the GT generators (Tab. 1 in the main paper). Specifically, SwAG produces SoTA performance on the discovery task. An exception is MAE which performs poorly compared to the other models. This can be attributed to its necessity for finetuning with a non-linear head before transferring to downstream tasks.



\section{Optimization details}
Our supervised finetuned baselines are complemented by a linear classification head (at the top of the backbone) that matches the dimension of the number of IDs and/or categories. By utilizing the attributed ID and/or category information of each gallery item, we minimize the categorical cross-entropy loss for $50$ epochs. It is used with Adam optimizer, with weight decay of $w=5\cdot10^{-5}$, and no momentum. The size of the mini-batches is set to be $64$. 

\begin{table}[h]
  \begin{center}
  \begin{tabular}{lcccccc}
    \toprule
\multirow{1}{*}{} & \multicolumn{1}{c}{HR} & \multicolumn{1}{c}{MRR} &	\multicolumn{2}{c}{ROC-AUC} & \multicolumn{2}{c}{PR-AUC} 	\\
\cmidrule(r){2-2}
\cmidrule(r){3-3}
\cmidrule(r){4-5}
\cmidrule(r){6-7}
& @5 	& @5 	&   Micro &  Macro &  Micro &  Macro \\
\cmidrule(r){2-2}
\cmidrule(r){3-3}
\cmidrule(r){4-4}
\cmidrule(r){5-5}
\cmidrule(r){6-6}
\cmidrule(r){7-7}
RN50 & 26.5 & 30.9 & 67.4 & 80.7 & 93.4 & 93.6 \\
MoCo & 32.9 & 39.0 & 63.8 & 78.7 & 92.5 & 93.0 \\
SwAV & 41.0 & \textbf{48.0} & 63.9 & 79.0 & 92.6 & 93.1 \\
MAE & 13.3 & 17.0 & 59.6 & 77.1 & 91.1 & 92.5 \\
CN & 24.6 & 29.0 & 66.1 & 79.7& 93.1  & 93.4 \\
B-16 & 34.3 & 40.5 & 68.9 & 81.7 & 94.0 & 94.1 \\
NS & 24.0 & 28.6 & 69.0 & 82.2 & 93.8 & 94.2 \\
SwAG& \textbf{40.3} & 47.3 & \textbf{75.4} & \textbf{85.0} & \textbf{95.4} & \textbf{95.2} \\
\bottomrule
  \end{tabular}
  \end{center}
\caption{Closed-catalog discovery performance.}
\label{tab:discovery_extra}
\end{table}

\begin{table}[ht]
  \begin{center}
  \begin{tabular}{lcccccc}
    \toprule
\multirow{1}{*}{} & \multicolumn{1}{c}{HR} & \multicolumn{1}{c}{MRR} &	\multicolumn{2}{c}{ROC-AUC} & \multicolumn{2}{c}{PR-AUC} 	\\
\cmidrule(r){2-2}
\cmidrule(r){3-3}
\cmidrule(r){4-5}
\cmidrule(r){6-7}
& @5 	& @5 	&   Micro &  Macro &  Micro &  Macro \\
\cmidrule(r){2-2}
\cmidrule(r){3-3}
\cmidrule(r){4-4}
\cmidrule(r){5-5}
\cmidrule(r){6-6}
\cmidrule(r){7-7}
RN50 & 6.0 & 6.7 & 64.4 & 71.1 & 54.3 & 66.0 \\
MoCo & 8.5 & 9.8 & 62.1 & 67.3 & 52.0 & 63.1 \\
SwAV & 8.3 & 9.6 & 62.4 & 68.1 & 52.1 & 63.5 \\
MAE & 0.6 & 0.7 & 47.4 & 51.7 & 39.8 & 51.7 \\
CN & 5.5 & 6.3 & 65.9 & 71.9 & 56.1 & 66.7 \\
B-16 & 8.0 & 9.4 & 67.8 & 74.3 & 58.8 & 69.8 \\
NS & 9.0 & 10.4 & 70.8 & 76.5 & 62.0 & 70.5 \\
SwAG& \textbf{10.0} & \textbf{11.1} & \textbf{72.4} & \textbf{78.5} & \textbf{63.4} & \textbf{72.7} \\
\bottomrule
  \end{tabular}
\end{center}
  \caption{Image in the wild discovery performance.}
  \label{tab:wild_extra}
\end{table}

\end{document}

%% file: 01_abstract.tex
\gls{Task} is an important task with broad e-commerce applications.
Given an image of a certain object, the goal of \gls{Task} is to retrieve images of \emph{different} objects with high perceptual visual similarity. 
Although being a highly addressed problem, the evaluation of proposed methods for \gls{Task} is often based on a proxy of an identification-retrieval task, evaluating the ability of a model to retrieve \emph{different} images of the \emph{same} object.
We posit that evaluating \gls{Task} methods based on identification tasks is limited, and faithful evaluation must rely on expert annotations. 
In this paper, we introduce the first large-scale fashion visual similarity benchmark dataset, consisting of more than 110K expert-annotated image pairs. Besides this major contribution, we share insight from the challenges we faced while curating this dataset.
Based on these insights, we propose a novel and efficient labeling procedure that can be applied to any dataset. Our analysis examines its limitations and inductive biases, and based on these findings, we propose metrics to mitigate those limitations. Though our primary focus lies on visual similarity, the methodologies we present have broader applications for discovering and evaluating perceptual similarity across various domains.

%% file: 02_intro.tex
Visual similarity measures the perceptual agreement between two objects based on their visual appearance \cite{palmer1978structural}. 
Two objects can be similar or dissimilar based on their color, shape, size, pattern, utility, and more. In fact, all of these factors and many others take part in determining the degree of visual similarity between two objects with varying importance. 
Therefore, defining the perceived visual similarity based on these factors is challenging. Nonetheless, learning visual similarities is a key building block for many practical utilities such as search, recommendations, etc.

\begin{figure}[t]
    \centering
    \includegraphics[scale=1.2]{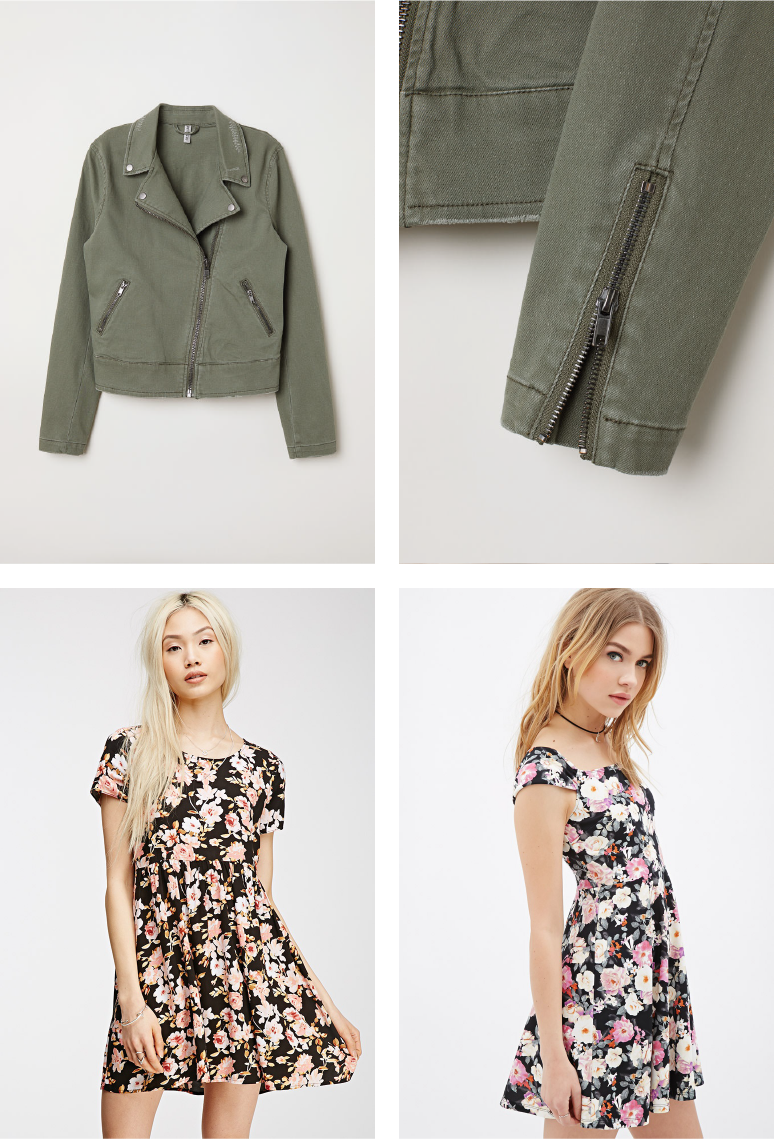}
	\caption{Disagreements between visual similarity and identification. \textit{Top:} Two images associated with the same object, exhibiting low visual perceptual similarity. \textit{Bottom:} Two images of two different objects, exhibiting high visual perceptual similarity.}\vspace{-6mm}
	\label{fig:similarity_vs_identity}
\end{figure}

Most existing methods for \gls{Task} are based on an identification retrieval task - given a query image of an object, the identification task deals with retrieving images of an \emph{identical} object taken under various conditions, such as different imaging distances, viewing angles, illuminations, backgrounds, and weather conditions e.g., \cite{zhai2018classification,wieczorek2020strong}. In fact, performance evaluation of image retrieval tasks is commonly gauged on identification-based classification metrics.

Identification and verification tasks~\cite{taigman2014deepface,barkan2013fast,dehak2010front,aronowitz2012efficient,barkan2013diffusion,snyder2018x} are in fact highly related to \gls{Task}, as any object is most similar to itself. Nevertheless, these two tasks are not the same (see \cref{fig:similarity_vs_identity}). Moreover, theoretically, a model that was trained for identification can obtain perfect results on the identification metrics, by retrieving other images of the same product followed by images of completely dissimilar products. 

An additional difficulty with learning identification as a proxy to similarity is simply the fact that multiple images of the same product are often unavailable. Furthermore, even when multiple images of the same product do exist, the images often do not conform with visual similarity, as illustrated in the upper row in \cref{fig:similarity_vs_identity}. 

In order to mitigate these difficulties, auxiliary information can be utilized. For example, tags~\cite{naka2022fashion,barkan2019cb2cf,barkan2020cold,barkan2021representation,barkan2020bayesian}, metadata~\cite{barkan2021cold,barkan2021cold2,malkiel2020recobert,malkiel2022metricbert,ginzburg2021self}, collaborative-filtering information~\cite{he2016vbpr,barkan2016item2vec,huang2015cross}, or explicit compatibility data from users~\cite{han2017learning}, were all employed in order to learn better visual similarities. However, such auxiliary information is often unavailable, and even when it is, it may not be a faithful proxy for perceived similarities. 

Ultimately we acknowledge that any proxy approach has its limitations and that the only faithful evaluation of visual perceptual similarity must rely on the annotations of human domain experts. However, employing such experts is time-consuming and expensive, and there was no publicly available dataset prior to this study. Thus, visual similarity models still rely on proxy evaluations, mostly the identification task, despite the aforementioned limitations. 

In this work, we address the challenge of discovering, curating, and evaluating visual similarity. We developed the Efficient Discovery of Similarities (EDS) method - a novel scheme for efficiently collecting feedback from human domain experts. By employing EDS we were able to label more than 110K image pairs which we release, as part of this paper, to serve as the first large-scale benchmark for evaluation of \gls{Task} models.

\textbf{Our contributions:} (1) We put a spotlight on the challenge of evaluating methods for \gls{Task}. We differentiate between the task of \gls{Task} and the identification task and stress the need for a true dataset of visual similarities. (2) We analyze the difficulty of naively labeling a dataset, and propose an efficient procedure for VSD, with proper evaluation metrics. The proposed method and metrics can be utilized for discovery and evaluation of perceptual similarity in other application domains. (3) Equipped with the proposed procedure, we curate the first large-scale visual similarity benchmark for the fashion domain, consisting of more than 110K labeled image pairs. This dataset enables true evaluations of perceived similarity and would help expedite further research in visual similarity and discovery. 
 (4) We provide an extensive evaluation comparing pretrained and finetuned models for both closed-catalog and wild queries. (5) Finally, we discuss and demonstrate the disadvantages of supervised methods for \gls{Task}.

%% file: 03_related.tex
In this paper, we address the challenge of evaluating methods for \gls{Task}. \gls{Task} is part of a fundamental computer vision task termed ``content-based image retrieval'' (CBIR)\footnote{https://en.wikipedia.org/wiki/Content-based\_image\_retrieval}, which involves ranking a catalog of images according to their similarity to a given query.

 \begin{figure}[t]
    \centering
    \includegraphics[width=0.42\textwidth]{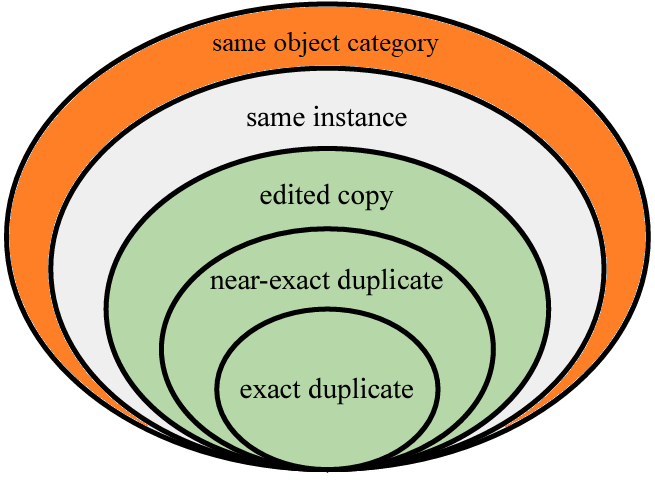} 
	\caption{Image-pair similarity defined using five tiers of granularity levels. The innermost tiers are the most restrictive and well-defined, and the outermost tiers are less restrictive and subject to the definition of ``instance'' and ``category''. The orange (green) area corresponds to the scope of \gls{Task} (ISC21).}\vspace{-5mm}
	\label{fig:similarity_tiers}
\end{figure}

Since similarity is subjective and depends on the application, evaluation of CBIR methods is a longstanding challenge \cite{eakins1999content}. This challenge was recently addressed in the Image Similarity Challenge at NeurIPS’21 (ISC21) \cite{douze20212021}. In \cite{douze20212021} the authors defined similarity using a 5-tiers granularity scheme. This scheme, taken from \cite{douze20212021}, is presented in \cref{fig:similarity_tiers}. The scope of \cite{douze20212021} corresponds to the green area in \cref{fig:similarity_tiers}. Since the objective of \gls{Task} is to retrieve images of \emph{different} objects with high visual perceptual similarity, the scope of \gls{Task} corresponds to the orange area.

Existing methods for \gls{Task} \cite{chopra2005learning,wang2014learning, razavian2016visual,schroff2015facenet,hadi2015buy,wu2017sampling,deng2019arcface} are often evaluated using the inner tier, marked by the gray color. The same practice is also common in the fashion domain. Existing methods for fashion retrieval tasks \cite{hadi2015buy, huang2015cross, liang2016clothes, corbiere2017leveraging,park2019study} are evaluated based on the ability to identify whether a pair of images correspond to the same instance or not. This evaluation is often carried out using publicly available datasets: DeepFashion \cite{deepfashion}, variations of Street2Shop \cite{liu2012street,hadi2015buy,wang2016matching}, and DARN \cite{huang2015cross}. In these datasets, each instance is associated with multiple images, taken under various conditions. For example, catalog images (in-shop) at different viewing angles, illuminations, and worn by different human models, or even images taken by customers (wild images). Since two different instances might exhibit high visual perceptual similarity, this evaluation is prone to false negatives. However, evaluation based on the category (the orange tier in \cref{fig:similarity_tiers}) is prone to false positives since two images of objects from the same category are not necessarily visually similar.

This trade-off can be mitigated by utilizing domain experts. In \cite{wang2014learning}, the authors proposed an annotations procedure based on popular text queries from the Google image search engine. However, this dataset is not oriented for the fashion domain, and their approach can not be applied to offline datasets.
In \cite{shankar2017deep}, the authors curated a fashion-expert annotated dataset for evaluation of their proposed \gls{Task} model, however, this dataset is proprietary.
To the best of our knowledge, the benchmark in this work would be the first large-scale benchmark for \gls{Task} in the fashion domain.

\textbf{Connection to classic Information-Retrieval.} 
The seminal Cranfield experiments were a series of information retrieval (IR) experiments conducted in the late 1950s and early 1960s at Cranfield University in England~\cite{cranfield}. These experiments were among the first large-scale, systematic evaluations of IR systems entailing the use of test collections of documents and queries, with the goal of evaluating the effectiveness of different IR systems in retrieving relevant documents in response to user queries. Relevance judgments were made by domain experts based on topical similarity in which documents and queries were carefully designed to represent the types of information retrieval tasks that might be encountered in real-world applications. Different IR systems were tested based on various measures such as precision, recall, and the F-measure. The Cranfield experiments helped to establish the importance of certain IR techniques, such as relevance feedback and query expansion. Retrieval test collections have also been employed by other researchers for system performance comparison using relevance judgments, as noted in~\cite{further_1,further_2}. 

Today, a comprehensive analysis of contemporary datasets is often impractical due to their large size. Consequently, \emph{pooling} techniques were developed for generating ground truth candidates. Pooling involves the selection of a fixed set of relevant documents for a given query, which are then used as the basis for computing evaluation metrics such as precision, recall, and the F-measure. 
Despite its effectiveness, pooling often results in incomplete labeled sets, potentially excluding relevant candidates that were not considered for the ground truth. This causes a bias that penalizes models that retrieve good candidates which were not shown to the annotators during labeling~\cite{zobel1998reliable}. 
To address this issue, evaluation metrics were developed that accommodate incomplete relevance assessments~\cite{metric1,metric2}. Our work relates to these works, as we address the issue of model bias resulting from incomplete labeling. 
To alleviate this bias, we introduce a metric to prioritize relative ranking over absolute ranking. Our approach builds upon similar techniques from the field of information retrieval that successfully handle incomplete information~\cite{buckley2004retrieval}. 

%% file: eds.tex
\section{Efficient discovery of similarities}
\label{sec:eds}
Let $\mathcal{D}$ be a dataset of images. Let $\mathcal{Q}\subset \mathcal{D}$ be a set of queries (for discussion, we assume $\mathcal{Q}$ is a subset of $\mathcal{D}$, while in the general case, $\mathcal{Q}$ and $\mathcal{D}$ can have partial or no overlap). Let $\mathcal{A}=\{(q,c)|q\in  \mathcal{Q}, c\in  \mathcal{D}_{-q}\}$ be the set of \emph{query-candidate} image pairs, where $ \mathcal{D}_{-q}:= \mathcal{D}\setminus\{q\}$. The general similarity labeling task is labeling the similarity for all image pairs in $\mathcal{A}$. Accordingly, the output of the labeling procedure is the set $\mathcal{Y}=\{y_{qc}|(q,c)\in \mathcal{A}\}$. Here, we assume binary labels, i.e., either the candidate image is similar to the query or not, and therefore, for all $(q,c)\in \mathcal{A}$ it holds that $y_{qc}\in\{0,1\}$. Thus, an image-pair $(q,c)$ is treated as \emph{positive} (\emph{negative}) if $y_{qc}=1$ ($y_{qc}=0$). 

In this work, we assume the pairs in $\mathcal{A}$ are labeled by a group of $E$ experts, and that each pair must be reviewed by \emph{all} experts. While the final decision regarding a label can be made according to various heuristics, in our labeling procedure, we followed a simple majority voting (where ties result in a negative label). Although more advanced methods are possible, they are outside the scope of this study.

In what follows, we consider two naive approaches for labeling image pairs in $\mathcal{A}$. These approaches illustrate the main challenge that motivated our proposed EDS method.

\subsection{Brute-force labeling}
\label{sec:bf-method}
The brute-force naive approach simply calls for human domain experts to label all pairs in $\mathcal{A}$. This method costs $O(| \mathcal{Q}|| \mathcal{D}|)$ labeling operations per expert, and becomes both prohibitive in time and expensive as $| \mathcal{D}|$ increases, hence impractical:
For example, consider a catalog of $| \mathcal{D}|=10,000$ candidates and $| \mathcal{Q}|=1,000$ queries, which results in ~$|\mathcal{A}|\approx10^7$ labeling operations per expert. Assuming the average time for labeling a pair is 15s, the total time to accomplish the labeling by a single expert that works 24/7 is more than 4.5 years. Of course, one could distribute the pairs in $\mathcal{A}$ among the $E$ experts to scale up the procedure by a factor of $E$, but then each partition will be annotated by a different expert which violates the requirement that each pair would be reviewed by all experts. 

\subsection{Random sample labeling}
\label{sec:random-method}
The main difficulty stems from the need to discover positive pairs in $\mathcal{A}$, which are extremely rare compared to the negative pairs. Denote $p$ as the fraction of positive pairs in $\mathcal{A}$. Then, the smaller the value of $p$, the longer it takes to discover a positive pair on average. Continuing the previous example, if we assume that the average number of positives per query is $k=10$, the total number of positives in $\mathcal{A}$ is $10,000$, and $p\approx0.001$ (the probability of sampling a positive pair at random). Therefore, the expected number of trials needed to obtain $h$ positives is $1,000h$. For example, we should expect the expert to annotate $100,000$ pairs to discover $100$ positives ($\sim$17 days of 24/7 work).

\subsection{The EDS method}
\label{sec:eds-method}
In this section, we present the EDS method that enables the efficient discovery of positive pairs. EDS utilizes a set of vision models to form a set $S$ of pairs \emph{suspected} to be positive. These suspects are then reviewed by experts that determine whether each suspected pair is indeed positive. The main assumption behind EDS is that the models can serve as a proxy similarity measure on the pairs in $\mathcal{A}$, and hence the produced set $S$ is likely to contain a fraction of true positives that are much higher than $p$ (Sec.~\ref{sec:random-method}). In what follows, we describe the method in detail, where we follow the same notations and setup as above. 

Let $M$ be a set of \gls{Task} models. We assume each model $m\in M$ is equipped with a similarity function $f_m:\mathcal{X}\times\mathcal{X}\rightarrow\mathbb{R}$, where $\mathcal{X}$ is the image domain. We define $R_m(q,c)$ as the (zero-based) rank of the image $c$ w.r.t. the query $q$, according to similarity scores produced by the application of $f_m$ to the pairs in $\mathcal{A}$. Then, we define the positive suspects set as $\mathcal{S}_k=\cup_{m\in M}\mathcal{S}_k^m$, with:
\begin{equation}
\label{eq:skm}
    \mathcal{S}_k^m=\{(q,c)|R_m(q,c)<k,(q,c)\in \mathcal{A}\}.
\end{equation}

Namely, $\mathcal{S}_k^m$ contains all $(q,c)$ for which $c$ is ranked in the top-$k$ images w.r.t. $q\in  \mathcal{Q}$, according to the model $m$, and $\mathcal{S}_k$ is the union of the $|M|$ sets. Note that $|\mathcal{S}_k|\leq|M|| \mathcal{Q}|k$ (and $|\mathcal{S}_k|=| \mathcal{Q}||M|k$ if $\mathcal{S}_k^m$ are disjoint). Therefore, the cost of a human domain expert to annotate all the pairs in $\mathcal{S}_k$ is $O(| \mathcal{Q}||M|k)$, whereas in the case of the brute-force method (Sec.~\ref{sec:bf-method}) the cost is $O(| \mathcal{Q}|| \mathcal{D}|)$. Finally, the output of EDS is an annotated set $\mathcal{Y}_k=\{y_{qc}|(q,c) \in \mathcal{S}_k\}$. 

In the common case, we get that $|M|k\ll| \mathcal{D}|$ hence EDS offers a significant reduction in labeling cost in large datasets. For example, in our experiments on the DF in-catalog dataset, we had $| \mathcal{D}|=52,712$, $| \mathcal{Q}|=2,000$, $|M|=6$, and $k=6$. In this case, the labeling procedure with EDS is $\sim$1,400x faster (and cheaper) than with the brute-force method. Yet, by employing EDS we introduce other challenges: (i) the number of annotated examples in the output set $\mathcal{A}_k$ is considerably less than the number of annotated examples in the case of the brute-force method $\mathcal{A}$ (ii) The annotated set $\mathcal{A}_k$ is biased towards the models that participated in the construction of $\mathcal{S}_k$. In Sec.~\ref{sec:eds-limitations}, we address these limitations in detail.

\subsubsection{Estimation of $p$}
\label{sec:p-estimation}
We define $p^m_k$ as the fraction of positive pairs in $\mathcal{S}_k^m$. As explained above, a prominent assumption in EDS is that the top-ranked candidates (by each model) produce positives with high probability, i.e., $p^m_k\gg p$, and therefore, $p_k\gg p$, where $p_k$ is the fraction of positive examples in $\mathcal{S}_k$.
For example, in the DF in-catalog dataset, we had $|\mathcal{S}_k|=54,170$ (after removing duplicate candidates suggested by different models). Among them, $45,920$ were labeled as positive pairs, resulting in $p_k = 0.848$. In addition, we can place a lower bound on the fraction of positives in $\mathcal{A}$ using $p_{LB}=\frac{45,920}{|\mathcal{A}|}=0.00045\leq p$.

One can estimate $p$ by random sampling of pairs from $\mathcal{A}$. Given a budget of $b$ labeling operations, we can sample $b$ pairs from $\mathcal{A}$, and annotate them. Then, we can compute the Maximum A-Posteriori (MAP) estimate for $p$ by using a uniform prior $U_p(p_{LB},1)$ together with a Binomial likelihood $B_p(a;b)$, where $a$ is the number of pairs that are annotated positive by the expert (out of the sample of $b$ pairs). One can show that the MAP estimate is given by:
\begin{equation}
\label{eq:p-estimator}
\hat{p}=\text{max}(p_{LB},\frac{a}{b}).
\end{equation}

Applying Chebyshev's inequality and setting $b = k/\epsilon$ yields:
\begin{equation}
\label{eq:p-estimator-error}
P(|\hat{p}-p| \ge \epsilon) \le \frac{p(1-p)}{b \epsilon^2} \le \frac{p}{b \epsilon^2}
\end{equation}
Assuming $p < \epsilon \ll 1$ and we want to bound the error with probability $q$, we need to use $b = 1 / {\epsilon q}$.

In order to estimate the improvement obtained by EDS over the random sampling method in terms of positive discovery rate, we randomly sampled $b=2,000$ pairs, annotated them with the experts, and obtained $a=2$. Accordingly, for the random sampling method from Sec.~\ref{sec:random-method}, we estimate $p$ with $\hat{p}=\frac{2}{2,000}=0.001$. Therefore, we conclude that in the specific case of the DF dataset, even if the error in our estimate is by a factor of 8 from the true value of $p$, the positive discovery rate with our EDS method ($p_k=0.848$) is still 100x higher than with the random sampling method.

\subsection{EDS limitations}
\label{sec:eds-limitations}
The first limitation of EDS is that it does not recover all the positive labels. However, all methods that label only a subset of query-result pairs share this limitation. Hence, it cannot be avoided when 
annotating any real-world data, for which the brute-force approach is infeasible.

\textbf{Model bias}. A second limitation of EDS concerns the fact that $\mathcal{A}_k$ contains labels for the image pairs in $\mathcal{S}_k$. As a result, the examples in $\mathcal{S}_k$ are heavily biased towards the models in $M$ (recall these models participated in the creation of $\mathcal{S}_k$). The bias in the dataset will lead to a bias in standard information retrieval metrics (e.g., Hit-Rate, Mean Reciprocal Rank, Normalized Discount Cumulative Gain, etc. \cite{tamm2021quality}). This bias can be especially prominent if $\mathcal{A}_k$ is used to evaluate the performance of any model $m' \notin M$ that did not participate in the creation of $\mathcal{S}_k$. To better understand why, let $\mathcal{S}_k^{m'}$ be the set of image pairs suggested by $m'$ (as defined in Eq.~\ref{eq:skm}), and assume that (1) $\mathcal{S}_k^{m'} \cap \mathcal{S}_k=\emptyset$, (2) all image pairs in $\mathcal{S}_k^{m'}$ would be assigned positive labels if they were to undergo expert evaluation, and (3) Hit-Rate is used for evaluation. Since the evaluation set is $\mathcal{S}_k$, the top-$k$ suggestions produced by $m'$ won't have labels. When computing the Hit-Rate we can either consider these suggestions negative or skip them, but both options lead to an unfair evaluation.
A naive solution for the bias problem would be to run the labeling procedure for each newly added model.
However, this approach is impractical, as it does not scale, and requires access to the same experts that labeled the original models.
In the next section, we propose metrics that mitigate the bias and enable the utilization of $\mathcal{S}_k$ for a fair evaluation of any model $m' \notin M$.

\section{Effective evaluation measures}
\label{sec:effective-eval}
In this section, we propose methods and metrics for fair evaluation w.r.t. the models that are not in $M$. To this end, we propose the \textit{area under the receiver operating curve} (ROC-AUC) as an evaluation measure that quantifies the ability of the model to rank positive pairs higher than negative pairs, regardless of their absolute rank. Specifically, ROC-AUC measures the probability that a random positive is ranked higher than a random negative. In what follows, we explain how to compute the ROC-AUC metric in the context of our work.

Let $\mathcal{P}_q=\{c|y_{qc}=1\}$ be the set of candidates labeled as positives w.r.t. the query $q$. Let $\mathcal{N}_q=\{c|y_{qc}=0\}$ be the set of candidates considered as negatives. Note that we consider negative candidates that were explicitly labeled as negatives. Furthermore, these negatives are ranked at the top-$k$, and therefore can be treated as $hard$ negatives that are more challenging (at least to the model that suggests them). However, it is also possible to extend $\mathcal{N}_q$ with a random sample of candidates that do not belong to $\mathcal{N}_q$. While the labels for these candidates are unknown, the probability of each randomly sampled candidate being a true negative is very high (recall that in DF our estimate for $p$ was 0.001). 

The ROC-AUC for the model $m$ on the query $q$ is:
\begin{multline}
\label{eq:auc}
\text{ROC-AUC}^m_q= \\
\frac{1}{|\mathcal{P}_q||\mathcal{N}_q|} 
\sum_{c\in \mathcal{P}_q}\sum_{c'\in \mathcal{N}_q} {\large{\mathbbm{1}}}{[R_m(q,c) < R_m(q,c')]}
\end{multline}
where $\mathbbm{1}[\cdot]$ is the indicator function. Given a set of queries $\mathcal{Q}$, the ROC-AUC for the model $m$ on $\mathcal{Q}$ is computed by:
\begin{equation}
\label{eq:avg_auc}
\text{ROC-AUC}^m_ \mathcal{Q}=\frac{1}{| \mathcal{D}|}\sum_{q\in  \mathcal{Q}}\text{ROC-AUC}^m_q
\end{equation}

The motivation for using ROC-AUC is as follows: given that we trust in the generated labels $\mathcal{A}_k$, these labels are treated as a faithful ground truth. Hence, a good model is expected to be able to differentiate between the positive and negative pairs, whether it belongs to $M$ or not. In contrast to metrics that focus on the absolute rank (e.g., HR, MRR, etc.), the ROC-AUC metric focuses on the relative rank (of the positive w.r.t. the negative). Therefore, the ROC-AUC should not favor models that participated in the ground truth generation procedure over models that did not. For example, consider the model $m'\in M$ discussed above. According to assumptions (1) and (2) from \cref{sec:eds-limitations}, all image pairs in $\mathcal{S}_k^{m'}$ would be assigned positive labels if were to go under expert evaluation.
Therefore, $m'$ will completely fail in the HR test, however in order to fail in the ROC-AUC test, $m'$ should rank negatives from $\mathcal{S}_k$ above the positive ones. In other words, the ROC-AUC metric is agnostic to the fact that the top-$k$ suggestions by $m'$ are not labeled, and will punish $m'$ only if it ranks negatives above positives.

%% file: egpaper_for_review.bbl
\begin{thebibliography}{10}\itemsep=-1pt

\bibitem{aronowitz2012efficient}
Hagai Aronowitz and Oren Barkan.
\newblock Efficient approximated i-vector extraction.
\newblock In {\em 2012 IEEE international conference on acoustics, speech and
  signal processing (ICASSP)}, pages 4789--4792. IEEE, 2012.

\bibitem{aucouturier2002music}
Jean-Julien Aucouturier, Francois Pachet, et~al.
\newblock Music similarity measures: What's the use?
\newblock In {\em Ismir}, volume~7, pages 339--340, 2002.

\bibitem{beit}
Hangbo Bao, Li Dong, and Furu Wei.
\newblock Beit: Bert pre-training of image transformers.
\newblock {\em arXiv preprint arXiv:2106.08254}, 2021.

\bibitem{barkan2017bayesian}
Oren Barkan.
\newblock Bayesian neural word embedding.
\newblock In {\em Proceedings of the AAAI Conference on Artificial
  Intelligence}, volume~31, 2017.

\bibitem{barkan2013diffusion}
Oren Barkan and Hagai Aronowitz.
\newblock Diffusion maps for plda-based speaker verification.
\newblock In {\em 2013 IEEE International Conference on Acoustics, Speech and
  Signal Processing}, pages 7639--7643. IEEE, 2013.

\bibitem{barkan2020cold}
Oren Barkan, Avi Caciularu, Idan Rejwan, Ori Katz, Jonathan Weill, Itzik
  Malkiel, and Noam Koenigstein.
\newblock Cold item recommendations via hierarchical item2vec.
\newblock In {\em 2020 IEEE International Conference on Data Mining (ICDM)},
  pages 912--917. IEEE, 2020.

\bibitem{barkan2021representation}
Oren Barkan, Avi Caciularu, Idan Rejwan, Ori Katz, Jonathan Weill, Itzik
  Malkiel, and Noam Koenigstein.
\newblock Representation learning via variational bayesian networks.
\newblock In {\em Proceedings of the 30th ACM International Conference on
  Information \& Knowledge Management}, pages 78--88, 2021.

\bibitem{barkan2021cold}
Oren Barkan, Roy Hirsch, Ori Katz, Avi Caciularu, Jonathan Weill, and Noam
  Koenigstein.
\newblock Cold item integration in deep hybrid recommenders via tunable
  stochastic gates.
\newblock In {\em 2021 IEEE International Conference on Data Mining (ICDM)},
  pages 994--999. IEEE, 2021.

\bibitem{barkan2021cold2}
Oren Barkan, Roy Hirsch, Ori Katz, Avi Caciularu, Yoni Weill, and Noam
  Koenigstein.
\newblock Cold start revisited: A deep hybrid recommender with cold-warm item
  harmonization.
\newblock In {\em ICASSP 2021-2021 IEEE International Conference on Acoustics,
  Speech and Signal Processing (ICASSP)}, pages 3260--3264. IEEE, 2021.

\bibitem{barkan2016item2vec}
Oren Barkan and Noam Koenigstein.
\newblock Item2vec: neural item embedding for collaborative filtering.
\newblock In {\em 2016 IEEE 26th International Workshop on Machine Learning for
  Signal Processing (MLSP)}, pages 1--6. IEEE, 2016.

\bibitem{barkan2019cb2cf}
Oren Barkan, Noam Koenigstein, Eylon Yogev, and Ori Katz.
\newblock Cb2cf: a neural multiview content-to-collaborative filtering model
  for completely cold item recommendations.
\newblock In {\em Proceedings of the 13th ACM Conference on Recommender
  Systems}, pages 228--236, 2019.

\bibitem{barkan2020scalable}
Oren Barkan, Noam Razin, Itzik Malkiel, Ori Katz, Avi Caciularu, and Noam
  Koenigstein.
\newblock Scalable attentive sentence pair modeling via distilled sentence
  embedding.
\newblock In {\em Proceedings of the AAAI Conference on Artificial
  Intelligence}, volume~34, pages 3235--3242, 2020.

\bibitem{barkan2020bayesian}
Oren Barkan, Idan Rejwan, Avi Caciularu, and Noam Koenigstein.
\newblock Bayesian hierarchical words representation learning.
\newblock {\em arXiv preprint arXiv:2004.07126}, 2020.

\bibitem{barkan2013fast}
Oren Barkan, Jonathan Weill, Lior Wolf, and Hagai Aronowitz.
\newblock Fast high dimensional vector multiplication face recognition.
\newblock In {\em Proceedings of the IEEE international conference on computer
  vision}, pages 1960--1967, 2013.

\bibitem{metric1}
Chris Buckley and Ellen~M Voorhees.
\newblock Retrieval evaluation with incomplete information.
\newblock In {\em Proceedings of the 27th annual international ACM SIGIR
  conference on Research and development in information retrieval}, pages
  25--32, 2004.

\bibitem{buckley2004retrieval}
Chris Buckley and Ellen~M Voorhees.
\newblock Retrieval evaluation with incomplete information.
\newblock In {\em Proceedings of the 27th annual international ACM SIGIR
  conference on Research and development in information retrieval}, pages
  25--32, 2004.

\bibitem{swav}
Mathilde Caron, Ishan Misra, Julien Mairal, Priya Goyal, Piotr Bojanowski, and
  Armand Joulin.
\newblock Unsupervised learning of visual features by contrasting cluster
  assignments.
\newblock {\em Advances in neural information processing systems},
  33:9912--9924, 2020.

\bibitem{dino}
Mathilde Caron, Hugo Touvron, Ishan Misra, Herv{\'e} J{\'e}gou, Julien Mairal,
  Piotr Bojanowski, and Armand Joulin.
\newblock Emerging properties in self-supervised vision transformers.
\newblock In {\em Proceedings of the IEEE/CVF International Conference on
  Computer Vision}, pages 9650--9660, 2021.

\bibitem{chopra2005learning}
Sumit Chopra, Raia Hadsell, and Yann LeCun.
\newblock Learning a similarity metric discriminatively, with application to
  face verification.
\newblock In {\em 2005 IEEE Computer Society Conference on Computer Vision and
  Pattern Recognition (CVPR'05)}, volume~1, pages 539--546. IEEE, 2005.

\bibitem{cranfield}
Cyril~W. Cleverdon.
\newblock The cranfield tests on index language devices.
\newblock 1997.

\bibitem{corbiere2017leveraging}
Charles Corbiere, Hedi Ben-Younes, Alexandre Ram{\'e}, and Charles Ollion.
\newblock Leveraging weakly annotated data for fashion image retrieval and
  label prediction.
\newblock In {\em Proceedings of the IEEE international conference on computer
  vision workshops}, pages 2268--2274, 2017.

\bibitem{pr}
Jesse Davis and Mark Goadrich.
\newblock The relationship between precision-recall and roc curves.
\newblock In {\em Proceedings of the 23rd international conference on Machine
  learning}, pages 233--240, 2006.

\bibitem{dehak2010front}
Najim Dehak, Patrick~J Kenny, R{\'e}da Dehak, Pierre Dumouchel, and Pierre
  Ouellet.
\newblock Front-end factor analysis for speaker verification.
\newblock {\em IEEE Transactions on Audio, Speech, and Language Processing},
  19(4):788--798, 2010.

\bibitem{deng2019arcface}
Jiankang Deng, Jia Guo, Niannan Xue, and Stefanos Zafeiriou.
\newblock Arcface: Additive angular margin loss for deep face recognition.
\newblock In {\em Proceedings of the IEEE/CVF conference on computer vision and
  pattern recognition}, pages 4690--4699, 2019.

\bibitem{further_1}
S.~F. Dierk.
\newblock The smart retrieval system: Experiments in automatic document
  processing — gerard salton, ed.
\newblock {\em IEEE Transactions on Professional Communication},
  PC-15(1):17--17, 1972.

\bibitem{vit}
Alexey Dosovitskiy, Lucas Beyer, Alexander Kolesnikov, Dirk Weissenborn,
  Xiaohua Zhai, Thomas Unterthiner, Mostafa Dehghani, Matthias Minderer, Georg
  Heigold, Sylvain Gelly, et~al.
\newblock An image is worth 16x16 words: Transformers for image recognition at
  scale.
\newblock {\em arXiv preprint arXiv:2010.11929}, 2020.

\bibitem{douze20212021}
Matthijs Douze, Giorgos Tolias, Ed Pizzi, Zo{\"e} Papakipos, Lowik Chanussot,
  Filip Radenovic, Tomas Jenicek, Maxim Maximov, Laura Leal-Taix{\'e}, Ismail
  Elezi, et~al.
\newblock The 2021 image similarity dataset and challenge.
\newblock {\em arXiv preprint arXiv:2106.09672}, 2021.

\bibitem{eakins1999content}
John~P Eakins and Margaret~E Graham.
\newblock Content-based image retrieval, a report to the jisc technology
  applications programme, 1999.

\bibitem{roc}
Tom Fawcett.
\newblock An introduction to roc analysis.
\newblock {\em Pattern recognition letters}, 27(8):861--874, 2006.

\bibitem{ginzburg2021self}
Dvir Ginzburg, Itzik Malkiel, Oren Barkan, Avi Caciularu, and Noam Koenigstein.
\newblock Self-supervised document similarity ranking via contextualized
  language models and hierarchical inference.
\newblock {\em arXiv preprint arXiv:2106.01186}, 2021.

\bibitem{hadi2015buy}
M Hadi~Kiapour, Xufeng Han, Svetlana Lazebnik, Alexander~C Berg, and Tamara~L
  Berg.
\newblock Where to buy it: Matching street clothing photos in online shops.
\newblock In {\em Proceedings of the IEEE international conference on computer
  vision}, pages 3343--3351, 2015.

\bibitem{han2017learning}
Xintong Han, Zuxuan Wu, Yu-Gang Jiang, and Larry~S Davis.
\newblock Learning fashion compatibility with bidirectional lstms.
\newblock In {\em Proceedings of the 25th ACM international conference on
  Multimedia}, pages 1078--1086, 2017.

\bibitem{mae}
Kaiming He, Xinlei Chen, Saining Xie, Yanghao Li, Piotr Doll{\'a}r, and Ross
  Girshick.
\newblock Masked autoencoders are scalable vision learners.
\newblock In {\em Proceedings of the IEEE/CVF Conference on Computer Vision and
  Pattern Recognition}, pages 16000--16009, 2022.

\bibitem{moco}
Kaiming He, Haoqi Fan, Yuxin Wu, Saining Xie, and Ross Girshick.
\newblock Momentum contrast for unsupervised visual representation learning.
\newblock In {\em Proceedings of the IEEE/CVF conference on computer vision and
  pattern recognition}, pages 9729--9738, 2020.

\bibitem{resnet}
Kaiming He, Xiangyu Zhang, Shaoqing Ren, and Jian Sun.
\newblock Deep residual learning for image recognition.
\newblock In {\em Proceedings of the IEEE conference on computer vision and
  pattern recognition}, pages 770--778, 2016.

\bibitem{he2016vbpr}
Ruining He and Julian McAuley.
\newblock Vbpr: visual bayesian personalized ranking from implicit feedback.
\newblock In {\em Proceedings of the AAAI conference on artificial
  intelligence}, volume~30, 2016.

\bibitem{huang2015cross}
Junshi Huang, Rogerio~S Feris, Qiang Chen, and Shuicheng Yan.
\newblock Cross-domain image retrieval with a dual attribute-aware ranking
  network.
\newblock In {\em Proceedings of the IEEE international conference on computer
  vision}, pages 1062--1070, 2015.

\bibitem{jiang2019smart}
Haoming Jiang, Pengcheng He, Weizhu Chen, Xiaodong Liu, Jianfeng Gao, and Tuo
  Zhao.
\newblock Smart: Robust and efficient fine-tuning for pre-trained natural
  language models through principled regularized optimization.
\newblock {\em arXiv preprint arXiv:1911.03437}, 2019.

\bibitem{knees2013survey}
Peter Knees and Markus Schedl.
\newblock A survey of music similarity and recommendation from music context
  data.
\newblock {\em ACM Transactions on Multimedia Computing, Communications, and
  Applications (TOMM)}, 10(1):1--21, 2013.

\bibitem{hr}
Walid Krichene and Steffen Rendle.
\newblock On sampled metrics for item recommendation.
\newblock {\em Communications of the ACM}, 65(7):75--83, 2022.

\bibitem{liang2016clothes}
Xiaodan Liang, Liang Lin, Wei Yang, Ping Luo, Junshi Huang, and Shuicheng Yan.
\newblock Clothes co-parsing via joint image segmentation and labeling with
  application to clothing retrieval.
\newblock {\em IEEE Transactions on Multimedia}, 18(6):1175--1186, 2016.

\bibitem{liu2012street}
Si Liu, Zheng Song, Meng Wang, Changsheng Xu, Hanqing Lu, and Shuicheng Yan.
\newblock Street-to-shop: Cross-scenario clothing retrieval via parts alignment
  and auxiliary set.
\newblock In {\em Proceedings of the 20th ACM international conference on
  Multimedia}, pages 1335--1336, 2012.

\bibitem{deepfashion}
Ziwei Liu, Ping Luo, Shi Qiu, Xiaogang Wang, and Xiaoou Tang.
\newblock Deepfashion: Powering robust clothes recognition and retrieval with
  rich annotations.
\newblock In {\em Proceedings of the IEEE conference on computer vision and
  pattern recognition}, pages 1096--1104, 2016.

\bibitem{convnet}
Zhuang Liu, Hanzi Mao, Chao-Yuan Wu, Christoph Feichtenhofer, Trevor Darrell,
  and Saining Xie.
\newblock A convnet for the 2020s.
\newblock In {\em Proceedings of the IEEE/CVF Conference on Computer Vision and
  Pattern Recognition}, pages 11976--11986, 2022.

\bibitem{malkiel2020recobert}
Itzik Malkiel, Oren Barkan, Avi Caciularu, Noam Razin, Ori Katz, and Noam
  Koenigstein.
\newblock Recobert: A catalog language model for text-based recommendations.
\newblock {\em arXiv preprint arXiv:2009.13292}, 2020.

\bibitem{malkiel2022metricbert}
Itzik Malkiel, Dvir Ginzburg, Oren Barkan, Avi Caciularu, Yoni Weill, and Noam
  Koenigstein.
\newblock Metricbert: Text representation learning via self-supervised triplet
  training.
\newblock In {\em ICASSP 2022-2022 IEEE International Conference on Acoustics,
  Speech and Signal Processing (ICASSP)}, pages 1--5. IEEE, 2022.

\bibitem{metric2}
Alistair Moffat, William Webber, and Justin Zobel.
\newblock Strategic system comparisons via targeted relevance judgments.
\newblock In {\em Proceedings of the 30th annual international ACM SIGIR
  conference on research and development in information retrieval}, pages
  375--382, 2007.

\bibitem{mueller2016siamese}
Jonas Mueller and Aditya Thyagarajan.
\newblock Siamese recurrent architectures for learning sentence similarity.
\newblock In {\em Proceedings of the AAAI conference on artificial
  intelligence}, volume~30, 2016.

\bibitem{naka2022fashion}
Rino Naka, Marie Katsurai, Keisuke Yanagi, and Ryosuke Goto.
\newblock Fashion style-aware embeddings for clothing image retrieval.
\newblock In {\em Proceedings of the 2022 International Conference on
  Multimedia Retrieval}, pages 49--53, 2022.

\bibitem{palmer1978structural}
Stephen~E Palmer.
\newblock Structural aspects of visual similarity.
\newblock {\em Memory \& Cognition}, 6(2):91--97, 1978.

\bibitem{pampalk2005improvements}
Elias Pampalk, Arthur Flexer, Gerhard Widmer, et~al.
\newblock Improvements of audio-based music similarity and genre classificaton.
\newblock In {\em ISMIR}, volume~5, pages 634--637. London, UK, 2005.

\bibitem{park2019study}
Sanghyuk Park, Minchul Shin, Sungho Ham, Seungkwon Choe, and Yoohoon Kang.
\newblock Study on fashion image retrieval methods for efficient fashion visual
  search.
\newblock In {\em Proceedings of the IEEE/CVF Conference on Computer Vision and
  Pattern Recognition Workshops}, pages 0--0, 2019.

\bibitem{mrr}
Dragomir~R Radev, Hong Qi, Harris Wu, and Weiguo Fan.
\newblock Evaluating web-based question answering systems.
\newblock In {\em LREC}. Citeseer, 2002.

\bibitem{clip}
Alec Radford, Jong~Wook Kim, Chris Hallacy, Aditya Ramesh, Gabriel Goh,
  Sandhini Agarwal, Girish Sastry, Amanda Askell, Pamela Mishkin, Jack Clark,
  et~al.
\newblock Learning transferable visual models from natural language
  supervision.
\newblock In {\em International Conference on Machine Learning}, pages
  8748--8763. PMLR, 2021.

\bibitem{razavian2016visual}
Ali~S Razavian, Josephine Sullivan, Stefan Carlsson, and Atsuto Maki.
\newblock Visual instance retrieval with deep convolutional networks.
\newblock {\em ITE Transactions on Media Technology and Applications},
  4(3):251--258, 2016.

\bibitem{imagenet21k}
Tal Ridnik, Emanuel Ben-Baruch, Asaf Noy, and Lihi Zelnik-Manor.
\newblock Imagenet-21k pretraining for the masses.
\newblock {\em arXiv preprint arXiv:2104.10972}, 2021.

\bibitem{schroff2015facenet}
Florian Schroff, Dmitry Kalenichenko, and James Philbin.
\newblock Facenet: A unified embedding for face recognition and clustering.
\newblock In {\em Proceedings of the IEEE conference on computer vision and
  pattern recognition}, pages 815--823, 2015.

\bibitem{shankar2017deep}
Devashish Shankar, Sujay Narumanchi, HA Ananya, Pramod Kompalli, and Krishnendu
  Chaudhury.
\newblock Deep learning based large scale visual recommendation and search for
  e-commerce.
\newblock {\em arXiv preprint arXiv:1703.02344}, 2017.

\bibitem{swag}
Mannat Singh, Laura Gustafson, Aaron Adcock, Vinicius de Freitas~Reis, Bugra
  Gedik, Raj~Prateek Kosaraju, Dhruv Mahajan, Ross Girshick, Piotr Doll{\'a}r,
  and Laurens Van Der~Maaten.
\newblock Revisiting weakly supervised pre-training of visual perception
  models.
\newblock In {\em Proceedings of the IEEE/CVF Conference on Computer Vision and
  Pattern Recognition}, pages 804--814, 2022.

\bibitem{snyder2018x}
David Snyder, Daniel Garcia-Romero, Gregory Sell, Daniel Povey, and Sanjeev
  Khudanpur.
\newblock X-vectors: Robust dnn embeddings for speaker recognition.
\newblock In {\em 2018 IEEE international conference on acoustics, speech and
  signal processing (ICASSP)}, pages 5329--5333. IEEE, 2018.

\bibitem{taigman2014deepface}
Yaniv Taigman, Ming Yang, Marc'Aurelio Ranzato, and Lior Wolf.
\newblock Deepface: Closing the gap to human-level performance in face
  verification.
\newblock In {\em Proceedings of the IEEE conference on computer vision and
  pattern recognition}, pages 1701--1708, 2014.

\bibitem{tamm2021quality}
Yan-Martin Tamm, Rinchin Damdinov, and Alexey Vasilev.
\newblock Quality metrics in recommender systems: Do we calculate metrics
  consistently?
\newblock In {\em Fifteenth ACM Conference on Recommender Systems}, pages
  708--713, 2021.

\bibitem{further_2}
Ellen~M Voorhees.
\newblock The philosophy of information retrieval evaluation.
\newblock In {\em Evaluation of Cross-Language Information Retrieval Systems:
  Second Workshop of the Cross-Language Evaluation Forum, CLEF 2001 Darmstadt,
  Germany, September 3--4, 2001 Revised Papers 2}, pages 355--370. Springer,
  2002.

\bibitem{wang2014learning}
Jiang Wang, Yang Song, Thomas Leung, Chuck Rosenberg, Jingbin Wang, James
  Philbin, Bo Chen, and Ying Wu.
\newblock Learning fine-grained image similarity with deep ranking.
\newblock In {\em Proceedings of the IEEE conference on computer vision and
  pattern recognition}, pages 1386--1393, 2014.

\bibitem{wang2016matching}
Xi Wang, Zhenfeng Sun, Wenqiang Zhang, Yu Zhou, and Yu-Gang Jiang.
\newblock Matching user photos to online products with robust deep features.
\newblock In {\em Proceedings of the 2016 ACM on international conference on
  multimedia retrieval}, pages 7--14, 2016.

\bibitem{wieczorek2020strong}
Mikolaj Wieczorek, Andrzej Michalowski, Anna Wroblewska, and Jacek Dabrowski.
\newblock A strong baseline for fashion retrieval with person re-identification
  models.
\newblock In {\em International Conference on Neural Information Processing},
  pages 294--301. Springer, 2020.

\bibitem{wu2017sampling}
Chao-Yuan Wu, R Manmatha, Alexander~J Smola, and Philipp Krahenbuhl.
\newblock Sampling matters in deep embedding learning.
\newblock In {\em Proceedings of the IEEE international conference on computer
  vision}, pages 2840--2848, 2017.

\bibitem{noisystud}
Qizhe Xie, Minh-Thang Luong, Eduard Hovy, and Quoc~V Le.
\newblock Self-training with noisy student improves imagenet classification.
\newblock In {\em Proceedings of the IEEE/CVF conference on computer vision and
  pattern recognition}, pages 10687--10698, 2020.

\bibitem{zhai2018classification}
Andrew Zhai and Hao-Yu Wu.
\newblock Classification is a strong baseline for deep metric learning.
\newblock {\em arXiv preprint arXiv:1811.12649}, 2018.

\bibitem{zobel1998reliable}
Justin Zobel.
\newblock How reliable are the results of large-scale information retrieval
  experiments?
\newblock In {\em Proceedings of the 21st annual international ACM SIGIR
  conference on Research and development in information retrieval}, pages
  307--314, 1998.

\end{thebibliography}
